\pdfoutput=1

\documentclass[11pt]{article}

\usepackage[]{acl}

\usepackage{times}
\usepackage{latexsym}

\usepackage[T1]{fontenc}

\usepackage[utf8]{inputenc}

\usepackage{microtype}

\usepackage{inconsolata}

\usepackage{graphicx}
\usepackage{hyperref}
\usepackage{booktabs}
\usepackage{multirow}
\usepackage{tabularx}
\usepackage{adjustbox}
\usepackage{makecell}
\usepackage{caption}
\usepackage{subcaption}
\usepackage{amsmath}
\DeclareMathOperator*{\argmax}{arg\,max}

\title{Contextual Refinement of Translations: Large Language Models for Sentence and Document-Level Post-Editing}

\author{Sai Koneru$^{1}$,
  Miriam Exel$^{2}$,
  Matthias Huck$^{2}$, \textnormal{and}
  Jan Niehues$^{1}$ \\
  $^{1}$ Karlsruhe Institute of Technology \\
  $^{2}$ SAP SE, Dietmar-Hopp-Allee 16, 69190 Walldorf, Germany \\
  \texttt{\{sai.koneru, jan.niehues\}@kit.edu} \\
  \texttt{\{miriam.exel, matthias.huck\}@sap.com}}

\begin{document}
\maketitle
\begin{abstract}
Large language models (LLMs) have demonstrated considerable success in various natural language processing tasks, but they have yet to attain state-of-the-art performance in Neural Machine Translation (NMT). Nevertheless, their significant performance in tasks demanding a broad understanding and contextual processing shows their potential for translation. To exploit these abilities, we investigate using LLMs for MT and explore recent parameter-efficient fine-tuning techniques. Surprisingly, our initial experiments find that fine-tuning for translation purposes even led to performance degradation compared to in-context-learning. To overcome this, we propose an alternative approach: adapting LLMs as Automatic Post-Editors (APE) rather than direct translators. Building on ability of the LLM to handle long sequences, we also propose extending our approach to document-level translation. We show that leveraging Low-Rank-Adapter fine-tuning for APE can yield significant improvements across both sentence and document-level metrics while generalizing to out-of-domain data. Most notably, we achieve a state-of-the-art accuracy rate of 88.7\% on the ContraPro test set, which specifically assesses the model's ability to resolve pronoun ambiguities when translating from English to German. Lastly, during manual post-editing for document-level translation, the source sentences are iteratively annotated which can be used to refine further translations in the document. Here, we demonstrate that leveraging human corrections can significantly reduce the number of edits required for subsequent translations.


\end{abstract}

\section{Introduction}

\begin{figure}[!t]
\includegraphics[width=0.5\textwidth]{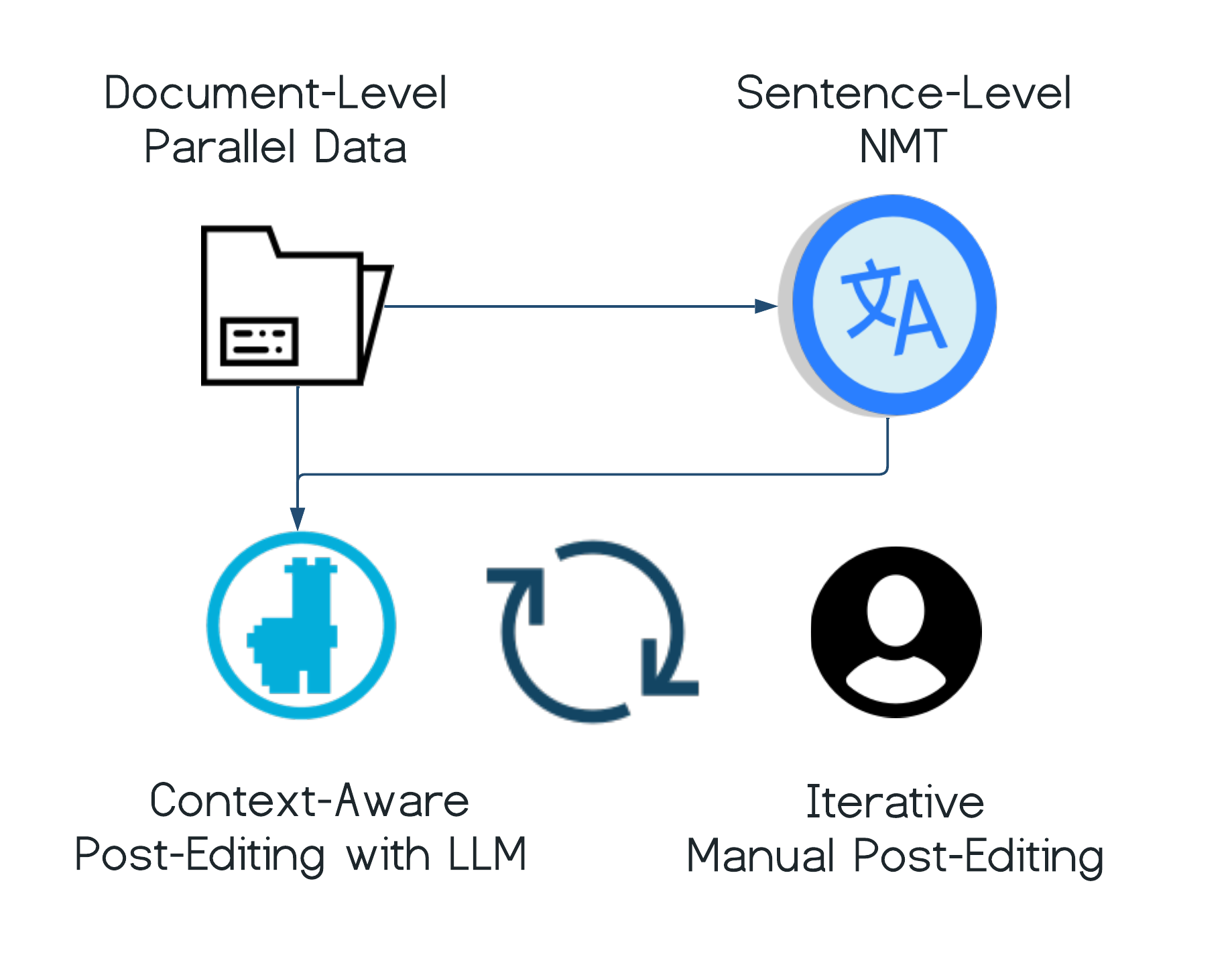}
 \caption{\textbf{Iterative Manual Post-Editing}: For manual PE, the annotator supplies the gold target context iteratively. The LLM then utilizes this gold target context for generating translations at document-level.}
\label{fig:approach}
\end{figure}

Large Language Models (LLMs) are currently being explored for many Natural Language Processing tasks such as Question Answering and Dialogue Applications \citep{touvron2023llama,anil2023palm,thoppilan2022lamda,tan2023evaluation}. Moreover, they are even shown to achieve or surpass state-of-the-art performance based on traditional methods. This achievement demonstrates their ability to possess general understanding and process long inputs. Given these strengths, LLMs might also be suitable for Machine Translation (MT) as many of them are also inherently multilingual from being trained on the web.

However, LLMs for MT still remain an under-explored area of research. While there are initial works on using LLMs for MT at both sentence and document-level \citep{vilar2022prompting, hendy2023good,wang2023document,zhang2023prompting}, the performance still lags behind the current state-of-the-art Neural MT (NMT) methods \citep{kocmi2023findings}.

It is worth noting that these methods mainly employ In-Context-Learning (ICL) \citep{brown2020language} as fine-tuning these models often requires a significant amount of computational resources. Hence, it might be a possible barrier to optimally adapt the LLMs for MT. 

Parameter-efficient techniques for fine-tuning such as Low-Rank Adapters (LoRA) \citep{hu2021LoRA,dettmers2023qLoRA} were recently proposed to overcome large computational requirements. This enables a new adaptation process for LLMs. However, it is still unclear whether these techniques are sufficient for successful adaptation and better generalization.

In this work, we investigate exploiting LLMs for MT at both sentence and document-levels. We initially experiment using ICL and parameter-efficient fine-tuning techniques to use LLM for MT and find that adding adapters alone is insufficient and may even lead to degradation (Section \ref{penecessary}). To mitigate this but still exploit the strengths of LLMs, we propose to adapt them as Automatic Post-Editors (APE) correcting NMT systems hypothesis rather than direct translators. 

This approach offers several advantages. It introduces modularity, allowing state-of-the-art or customized NMT techniques to be applied independently, followed by improvements made using the LLM. Additionally, LLMs can refine the sentence-level NMT systems output and generate consistent and coherent text using their ability to generate fluent and long documents. 

The cascaded system of NMT and LLM offering modularity can also enable the integration of human feedback. By feeding the LLM human-corrected translations from previous sentences in the document, we show that it can leverage this feedback to improve the current sentence's translation. This process can be applied iteratively in practice, sentence by sentence, as the annotator progressively corrects the translated document (see Figure \ref{fig:approach} for Iterative Manual Post-Editing).

We summarize our main findings and contributions below:
\begin{itemize}
    \item \textbf{Effective Combination of NMT and LLM}: Our sentence-level LLM APE demonstrates a successful fusion of knowledge from NMT systems and LLMs, leading to substantial enhancements in translation quality. Importantly, we observe that the LLM APE exhibits robustness and can adeptly correct NMT systems, even for test sets from different domains that it was not explicitly trained on.
    \item \textbf{Extension to Document-level Post-Editing}: We extend our approach to document-level APE and observe significant improvements in both sentence and document-level translation metrics. Notably, we achieve a state-of-the-art accuracy of 88.7\% on the ContraPro English to German test set, underscoring the effectiveness of APE.
    \item \textbf{Iterative Manual Post-Editing}: We introduce a promising use-case scenario for iterative post-editing (as depicted in Figure \ref{fig:approach}). We show that providing gold target context significantly enhances the remaining translation quality, both at the sentence and document levels, as indicated by various metrics.
\end{itemize}

\section{Approach: Adapting LLM for APE}
 LLMs may not be as proficient translators as state-of-the-art NMT systems due to no explicit training with large amounts of parallel data. However, LLMs being trained on the web containing data from several domains, possess general knowledge that is lacking in a NMT model. Moreover, they are capable of processing significantly longer inputs compared to a standard sentence-level NMT. Given their strengths of knowledge and ability to process long sequences, we propose to use them for APE for both sentence and document-levels. Hence, we first generate translations with NMT and then perform APE with LLMs. Our approach combines the translation capacity of NMT with fluency and understanding of LLMs. 

We use a technique similar to \citet{niehues-etal-2016-pre}, which combines Phrase-Based and NMT models. We extend this approach to incorporate LLMs for APE at both sentence and document-levels. We first explain the complete pipeline of our system at both levels of text representation. Then, we describe how we train and create the data for each step in our setup. Finally, we explain how human feedback can be easily integrated into our cascaded approach during manual PE.

\subsection{Pipeline}
Given a source sentence $s$, we use an NMT model to generate an initial translation $h_{NMT}$. Then for our APE model, we do not provide the $h_{NMT}$ alone as it cannot distinguish when the hypothesis from NMT is severely mistranslated but still fluent. Hence, we feed the source sentence and the initial translation to LLM and generate a refined hypothesis $h_{LLM}$:


\begin{align}
    h_{NMT} &= \argmax \log p_{\theta_{NMT}}(s) \\
    h_{LLM} &= \argmax \log p_{\theta_{LLM}}(s,h_{NMT})
\end{align}

where $\theta_{NMT}$ and $\theta_{LLM}$ are models trained for translation and APE. 

For APE at document-level, we extend the above formulation to process a sequence of sentences. Consider a document $\mathcal{D}$ with $n$ source sentences $s^{i}$ where $i$ ranges from $1$ to $n$. We first use the NMT model to generate each translation in isolation at sentence-level. Let them be denoted as $h_{NMT}^{i}$. Then, we perform APE using the sequence of source and hypothesis sentences exploiting the LLM ability to process and use contexts. We denote the generated document translation as $h_{LLM}^{D}$:


\begin{align}
    h_{NMT}^{i} &= \argmax \log p_{\theta_{NMT}}(s^{i}) \; \forall i\in1..n \\
    h_{LLM}^{D} &= \argmax \log p_{\theta_{LLM}}(s^{D},h_{NMT}^{D})
\end{align}

where $s^{D}$ and $h_{NMT}^{D}$ are the source and sentence-level hypothesis sentences joined by a separator token to form a document.

\subsection{LLM Fine-tuning for APE}
We have an NMT and an LLM in our cascaded approach. For training our NMT model, we fine-tune it on available parallel data in a conventional fashion. We do not do any additional steps as our main motivation was to exploit LLMs for further enhancements. In the case of the LLM, we propose to go beyond ICL approaches and fine-tune them for maximum utility as described in the following.

\subsubsection{Training on MT Errors}
\label{sec:generate}
To further optimize the LLM for the task of APE, we propose to fine-tune them using Q-LoRA \citep{hu2021LoRA,dettmers2023qLoRA}. It is ideal to fine-tune the LLM by providing the source and hypothesis as input and predicting the corresponding post-edited reference. This needs data in the form of triples comprising the source, initial hypothesis, and reference. To simulate real test conditions, we need the initial hypothesis to be consisting of the errors generated from the NMT model we plan to use.

To achieve this, we follow these steps:

\begin{enumerate}
    \item We partition the training data into two halves.
    \item We train two separate NMT models, one for each half of the data.
    \item We utilize the model trained on the first half to perform inference on the second half, and vice versa.
\end{enumerate}

This process yields the same quantity of instances as the original training set, with initial hypotheses that exhibit errors typical of the NMT system we plan to improve on. Subsequently, we format this data into a prompt template, as described in Appendix \ref{promptsentpe} or \ref{promptdocpe}, depending on the level of representation. Then, we employ Q-LoRA for fine-tuning\footnote{Training details can be found in Appendix \ref{sec:train_details}} our LLM. For document-level APE (Doc APE), we simply split into non-overlapping chunks according to a chunk limit of source tokens and create our training data.

\subsubsection{Inference}
In our setup, the level of granularity can be a sentence or a document. For sentence-level APE (Sent APE), the process during decoding is straightforward. We generate an initial translation and feed it to the adapted LLM for our final refined hypothesis.

In the context of document translation, decoding poses a more intricate challenge compared to sentence-level. Decisions must be made regarding the direction of context for each source sentence, whether it should be drawn from the left, right, or both sides. In our work, we explore the following strategies:

\textbf{Chunk-Based}: This is a straightforward approach where we employ the same method used to create our training data. We create non-overlapping chunks and translate them individually. In this setup, it's possible that some sentences may lack left or right context. Note that if the number of sentences in the hypothesis doesn't match the source, we replace them with the sentence-level $\triangle$ LM outputs exploiting the modularity of the cascaded approach. We've observed that this situation occurs infrequently, with at most 30 sentences, and thus for rare instances.

\textbf{Batched Sliding Window}: We translate the document using a sliding window approach with a payload, as described in \citet{post2023escaping}. We append the sentence we intend to translate with as much preceding source context as possible, following our chunk limit. Then, we translate the entire chunk, including the context (\textit{Payload}), and extract the last sentence using the separator token.

\textbf{Continuous Sliding Window}: Similar to the previous approach above, we append the left source context according to the chunk limit for translation. However, the key distinction here lies in not regenerating the target context at every step. When translating a sentence, we force-decode the translation of the previous sentences that are the target context in the next step. Hence, at each step, only one sentence is translated into the target language, which is then used for forced decoding in the subsequent step to provide the target context (Referred to as \textit{Sequential Decoding} in \citet{herold-ney-2023-improving}). 

\subsection{Integrating Manual Feedback}
Consider the case of manual PE where the annotator edits each sentence in the document. Here, we have access to the human-corrected target context that can be used to refine future translations. 

We propose to integrate this information into our APE system. By iteratively feeding the model human-corrected contextual information from preceding sentences and appending it to the prompt, we condition its subsequent translations on this expert knowledge. This modular approach enables straightforward integration of human input without requiring additional training.  

\begin{table*}[!ht]
\centering
\begin{tabular}{@{}c|ccccc@{}}
\toprule
\multirow{2}{*}{Dataset} & \multirow{2}{*}{Sentences} & \multirow{2}{*}{Documents} & \multicolumn{3}{c}{MuDA Tags}           \\ \cmidrule(l){4-6} 
                         &                            &                            & Pronouns & Formality & Lexical Cohesion \\ \midrule
MuST-C V3 Train          & 261.4K                     & 2.5K                       & 28K      & 82K       & 86K              \\
MuST-C V3 Test           & 3637                       & 14                         & 332      & 1127      & 1268             \\
WMT 21 News              & 1002                       & 68                         & 42       & 145       & 381              \\
ACL Dev                  & 468                        & 5                          & 12       & 38        & 478              \\ \bottomrule
\end{tabular}
\caption{Statistics of our training and test data sets. We report the number of sentences and documents along with the total tag occurrences annotated by the MuDA tagger.}
\label{tab:datastat}
\end{table*}

\section{Experimental Setup}
\textbf{Models:} In our proposed approach, we have a sentence-level NMT system that generates an initial hypothesis and an LLM which then improves it. Nonetheless, we want a strong NMT model to assess the benefits of using LLMs. Therefore, we fine-tune the pre-trained DeltaLM\footnote{We use the \href{https://github.com/microsoft/unilm/tree/master/deltalm}{$\triangle$LM base} model with $360M$ parameters} \citep{ma2021deltalm} ($\triangle$ LM) for initializing our NMT sentence-level model (Refer to Appendix \ref{sec:train_details_deltalm} for more details). For LLM, we use the recently open-sourced \href{https://huggingface.co/meta-llama/Llama-2-13b-chat-hf}{Llama-2-13b-chat-hf} \citep{touvron2023llama} as it is instruction-finetuned and has reasonable compute and memory requirements when adapting with 4bit Q-LoRA \citep{dettmers2023qLoRA}.

During training (Refer to Appendix \ref{sec:train_details} for more details), we mask the loss on the prompt, which means that the LLM is exclusively trained to predict the reference given the source and hypothesis. 

\textbf{Datasets \& Metrics:} We primarily focus on translating talks from \textbf{English to German} at a document-level. This choice is based on the large availability of document-level parallel data, the current state of sentence-level NMT quality, and the necessity for contextual information in this translation direction.

For training our sentence NMT and post-editor LLMs, we utilize the MuST-C V3 Corpus \citep{di-gangi-etal-2019-must}. This corpus aligns well with our objectives, as it contains parallel data annotated with talk IDs for document-level translation.

For testing, we report results on three test sets. First, we select a subset of the training corpus with the most contextual phenomena in the speeches. This choice stems from the fact that the need for context is not always prevalent, and hence, standard tests may not suffice for document-level evaluation. To identify such contextual phenomena, we employ the MuDA tagger \citep{fernandes-etal-2023-translation}, which automatically tags words requiring contextual information in the training corpus. We select talks with the highest number of tags for each phenomenon, resulting in 14 talks in our test sets, addressing contextual occurrences related to pronouns, formality, and lexical cohesion. Then, we remove the selected talks from our training data and use them for testing.

To evaluate the robustness of our approach with out-of-domain data, we use the WMT21 News test set \citep{akhbardeh2021findings} and the ACL dev set from IWSLT23 \citep{salesky-etal-2023-evaluating}. Although the ACL data consists of talks, its content contains terminology and domains unlikely to be found in the training data. Furthermore, both test sets are annotated at the document level, aligning with our experimental setup.

An overview of the datasets is presented in Table \ref{tab:datastat}. Notably, the number of detected tags is relatively small compared to the total sentences, even when accounting for false positives in WMT and ACL test sets. Therefore, creating a custom test set was essential to reliably evaluate the usage of context along with sentence-level translation quality.

We also report scores on the ContraPro (EN $\rightarrow$ DE) \citep{muller-etal-2018-large} for a targeted evaluation of context usage in resolving pronoun ambiguities.

Regarding metrics, we employ a variety to assess the quality at both the sentence and document-levels. Our report includes BLEU \citep{papineni2002bleu}, ChrF2 \citep{popovic2016chrf} using SacreBLEU \citep{post-2018-call}, and COMET\footnote{Unbabel/wmt22-comet-da} \citep{rei-etal-2022-comet} scores for sentence-level evaluation. To gauge the quality of word prediction using contextual information at the document-level, we report Precision, Recall, and F1 scores for words detected by the MuDA tagger.

\section{Automatic Post-Editing is Necessary}
\label{penecessary}

\begin{table}[!ht]
\resizebox{\columnwidth}{!}{
\begin{tabular}{cccc}
\hline
Shot Size                                                      & BLEU  & ChrF2 & COMET  \\ \hline
Sent-level $\triangle$ LM                                            & \textbf{30.45} & \textbf{57.0} & \textbf{0.8179} \\ \hline
\begin{tabular}[c]{@{}c@{}}Llama2 ICL\\ (Random 2)\end{tabular} & 21.53 & 50.0 & 0.7795 \\ \hline

Llama2 MT                                                           & 28.92 & 55.9 & 0.7664 \\ \hline
Llama2 $0$-Shot APE                                                   & 27.26 & 53.9 & 0.8008 \\ \hline
\end{tabular}
}
\caption{ICL, LoRA fine-tuning for MT and $0$-Shot APE performance of the Llama2 on Must-C test set. We report BLEU, ChrF2 and COMET scores for each configuration (EN $\rightarrow$ DE) and highlight the best scores in \textbf{bold} for each metric. We report only best performing ICL configuration but provide results of 1-5 shots in the Appendix \ref{sec:llmicl}.}
\label{tab:baselines}
\end{table}

While APE with LLMs seems intuitively advantageous, our first step is to empirically evaluate it against several baselines, including alternatives like In-Context-Learning (ICL) \citep{brown2020language} and fine-tuning with LoRA. To justify the development of a cascaded system with added computational complexity, we assess the following configurations\footnote{Prompt templates for all the configurations described in Appendix \ref{sec:prompt}}:

\textbf{Sentence-Level NMT}: We fine-tune $\triangle$LM on the training data at sentence-level in conventional fashion.

\textbf{In-Context-Learning with Llama2}: We prompt the model according to \citet{vilar-etal-2023-prompting}, selecting examples randomly or similar to the current prompt. To find sentences more closely related to our source, we extract sentence embeddings from our training data using Sent-BERT \citep{reimers-gurevych-2019-sentence} and retrieve the nearest neighbors for efficiency using FAISS \citep{johnson2019billion} (\textit{Llama2 ICL}).

\textbf{LLama2 + Adapters}: Leveraging the recent advancements in efficient fine-tuning of LLMs, we fine-tune with adapters using LoRA \citep{hu2021LoRA} (Training and Hyper-Parameter Details can be found in Appendix \ref{sec:train_details}). Like the sentence-level NMT, we fine-tune it on all of our training data at the sentence level (\textit{Llama2 MT}).

\textbf{Zero-Shot Post-Editing with Llama2}: Finally, we consider the case of simple zero-shot PE to evaluate the in-built ability of the model to use the knowledge from another system and compare it to ICL where it acts as a direct translator.

Results for the above setups are reported in Table \ref{tab:baselines}. First and foremost, we observe that the sentence-level $\triangle$ LM achieves the highest scores across all metrics. This underscores the highly competitive performance of a dedicated NMT model with 360M parameters compared to a 13B LLM.

Furthermore, we find that in the case of ICL, the selection strategy is relatively unimportant, with both random and FAISS  performing similarly as indicated by the scores in Appendix \ref{sec:llmicl}. Additionally, increasing the number of exemplars in the prompt had a detrimental effect on our setup. Moreover, adapting with LoRA yields the highest BLEU and ChrF2 scores of 28.92 and 55.9 when compared to setups that rely solely on the LLM. However, it's worth noting that COMET scores decrease compared to ICL. These findings align with those of \citet{xu2023paradigm}, where fine-tuning LLMs on extensive parallel data led to higher scores in lexical metrics but degradation in COMET.



Zero-Shot APE beats ICL across metrics (COMET included), unlike LoRA, showing LLMs' innate post-editing ability. However, it falls short of sentence-level $\triangle$ LM. Therefore, we propose to train the adapter for APE rather than direct translators for exploiting LLMs.

\section{Llama2 as Sentence-Level Post Editors}

\begin{table}[!ht]
\resizebox{\columnwidth}{!}{
\begin{tabular}{@{}cccc@{}}
\toprule
Model                                                                        & BLEU           & ChrF2          & COMET           \\ \midrule
\multicolumn{4}{c}{\textit{\textbf{MuST-C V3}}}                                                                                 \\ \midrule
$\triangle$ LM                                                               & 30.45          & 57            & 0.8179          \\
Llama2 MT                                                                   & 28.92          & 55.9          & 0.7663          \\
Llama2 $0$-Shot APE                                                                   & 27.26          & 53.9          & 0.8009          \\
\begin{tabular}[c]{@{}c@{}}$\triangle$ LM + Llama2 Sent APE\end{tabular} & \textbf{31.71} & \textbf{58.3} & \textbf{0.833}  \\ \midrule
\multicolumn{4}{c}{\textit{\textbf{WMT 21 News}}}                                                                               \\ \midrule
$\triangle$ LM                                                               & 21.53          & 52.6          & 0.7911          \\
Llama2 MT                                                                   & 23.61          & 54.3          & 0.7931          \\
Llama2 $0$-Shot APE                                                                   & 21.44          & 52.0          & 0.7982          \\
\begin{tabular}[c]{@{}c@{}}$\triangle$ LM + Llama2 Sent APE\end{tabular} & \textbf{25.16} & \textbf{56}   & \textbf{0.8411} \\ \midrule
\multicolumn{4}{c}{\textit{\textbf{ACL Dev}}}                                                                                   \\ \midrule
$\triangle$ LM                                                               & 31.36          & 60.5          & 0.7945          \\
Llama2 MT                                                                   & 31.47          & 60.5          & 0.772           \\
Llama2 $0$-Shot APE                                                                   & 30.83          & 60.3          & 0.8028          \\
\begin{tabular}[c]{@{}c@{}}$\triangle$ LM + Llama2 Sent APE\end{tabular} & \textbf{36}    & \textbf{63.9} & \textbf{0.8321} \\ \bottomrule
\end{tabular}
}
\caption{Performance of Sent-Level Llama2 APE on test sets in and out of the domain. \textit{$\triangle$ LM + Llama2 Sent APE} denotes using the hypothesis of $\triangle$ LM as input for our adapted Sentence-Level Llama2 post editor. We report BLEU, ChrF2, and COMET scores for each approach (EN $\rightarrow$ DE) and highlight the best scores in \textbf{bold} for each metric.}
\label{tab:sentpe_baselines}
\end{table}

In this section, we evaluate the performance of improving sentence translations with APE. First, we discuss the results of improving translations generated only by $\triangle$ LM on in-domain test data. Then, we analyze the influence of moving away from our training conditions to assess the robustness of the model. We achieve this by first evaluating its performance on out-of-domain test sets and combining it with hypotheses generated by models other than $\triangle$LM.






\subsection{Improved Translation Quality with Sentence-Level APE}

We evaluate our Sentence-Level Llama2 APE and present the results in Table \ref{tab:sentpe_baselines}. To assess the utility of APE, we also report scores for the individual models, namely, the sentence-level $\triangle$ LM and the Llama2 fine-tuned with LoRA on the parallel data.

We see that post-editing the output of $\triangle$ LM with Llama2 outperforms other models across all metrics while fine-tuning for MT alone shows degradation. We hypothesize that this is primarily due to LLMs' internal knowledge and intrinsic ability to generate fluent sentences while lacking in translation capability. However, by providing initial translations to make the task easier, LLM improves the quality by a high margin.

\subsection{Generalizability to Out-Of-Domain Data}

From Table \ref{tab:sentpe_baselines}, we observe that the performance gains are more pronounced on the WMT21 News and ACL test sets compared to our MuST-C test set. The primary difference is that WMT and ACL fall outside the domain of the training data. Hence, we observe more significant improvements compared to our MuST-C test set. We gain by $1.26$ BLEU on MuST-C but up to $5.64$ and $3.63$ BLEU on the out-of-domain ACL and WMT test sets respectively. 

This scenario mirrors practical situations where a system encounters out-of-domain sentences and performs sub-optimally. By utilizing Llama2, containing a broader spectrum of "knowledge" (Illustrated in Table \ref{tab:pesentexample}), we demonstrate that it can significantly enhance translation quality.

\subsection{Generalizability to NMT Models}
Apart from improving $\triangle$ LM hypothesis, it is ideal if the APE with Llama2 can enhance translations of various NMT models. Therefore, to critically assess the generalization ability of the Sentence-Level Llama2 APE, we evaluate it on correcting hypotheses that were not generated from $\triangle$ LM and out-of-domain ACL dev set. For this purpose, we utilize the NLLB models\footnote{We perform inference with 8-bit quantization and achieve slightly lower scores than reported in the literature} \citep{costa2022no} and present the results in Table \ref{tab:sentpe_nllb}.

\begin{table}[!ht]
\resizebox{\columnwidth}{!}{
\begin{tabular}{@{}cccc@{}}
\toprule
Model                                                                                   & BLEU           & ChrF2           & COMET          \\ \midrule
\multicolumn{4}{c}{\textit{\textbf{ACL Dev}}}                                                                                              \\ \midrule
Llama2 MT                                                                              & 31.47          & 60.5           & 0.772          \\
NLLB 3.3B                                                                               & 43.01          & 69.7           & 0.8321         \\
\begin{tabular}[c]{@{}c@{}}NLLB 3.3B + Llama2 Sent APE\end{tabular}                 & 40.09          & 67.2           & 0.8372*        \\
NLLB 54B                                                                  & \textbf{45.82} & \textbf{71.56} & \textbf{0.844} \\
\begin{tabular}[c]{@{}c@{}}NLLB 54B + Llama2 Sent APE\end{tabular} & 40.91          & 67.8           & 0.8407         \\
 \bottomrule
\end{tabular}
}
\caption{Analyzing the robustness of the Llama2 Sentence-Level APE. \textit{NLLB X LLM + LLM Sent APE} denotes using the hypothesis of NLLB X as input for our adapted Sentence-Level LLM APE. Best scores for a test set are in \textbf{bold} for each approach. If the post-editor improves the hypothesis according to a metric, we denote it with \textbf{*}}
\label{tab:sentpe_nllb}
\end{table}

\begin{table*}[!ht]
\resizebox{2\columnwidth}{!}{
\begin{tabular}{@{}c|ccc|ccc|ccc|ccc@{}}
\toprule
\multirow{2}{*}{Approach}                                 & \multirow{2}{*}{BLEU} & \multirow{2}{*}{ChrF2} & \multirow{2}{*}{COMET} & \multicolumn{3}{c|}{Pronouns}                & \multicolumn{3}{c|}{Formality}                & \multicolumn{3}{c}{Lexical Cohesion}          \\ \cmidrule(l){5-13} 
                                                          &                       &                        &                        & Precision     & Recall       & F1            & Precision     & Recall        & F1            & Precision     & Recall        & F1            \\ \midrule
$\triangle$ LM                                                  & 30.45                 & 57                     & 0.8179                 & 0.65          & 0.76         & 0.70          & 0.68          & 0.70          & 0.69          & 0.60          & 0.74          & 0.67          \\ \midrule
$\triangle$ LM Doc2Doc                                              & 30.66                 & 57.7                   & 0.7481                 & 0.66          & 0.78         & 0.71          & 0.68          & 0.72          & 0.69          & 0.6           & 0.74          & 0.66          \\ \midrule
Llama2 MT                                                  & 28.92                 & 55.9                   & 0.7663                 & 0.66          & 0.77         & 0.71          & 0.67          & 0.71          & 0.69          & 0.61          & 0.76          & 0.68          \\ \midrule
Llama2 MT Doc2Doc                                               & 28.98                 & 56.1                   & 0.8221                 & 0.67          & 0.75         & 0.71          & 0.68          & 0.74          & 0.71          & 0.61          & 0.70          & 0.65          \\ \midrule
$\triangle$ LM + Llama2 SENT APE                                               & 31.71                 & 58.3                   & 0.8330*                 & 0.66          & 0.77         & 0.71          & 0.67          & 0.71          & 0.69          & 0.61          & 0.76          & 0.68*         \\ \midrule
\begin{tabular}[c]{@{}c@{}}$\triangle$ LM + Llama2 Doc APE\\ Chunk\end{tabular}                                               & 31.47                 & 58.4                   & 0.8306                 & 0.68          & 0.82         & 0.74          & 0.66          & 0.76          & 0.71          & 0.60          & 0.76          & 0.67         \\ \midrule
\begin{tabular}[c]{@{}c@{}}$\triangle$ LM + Llama2 Doc APE\\ Batched SW\end{tabular}   & 31.77                 & 58.9*                   & 0.8300                   & 0.68          & 0.83*        & 0.75*         & 0.67          & 0.77          & 0.72          & 0.61          & 0.77* & 0.68*         \\ \midrule
\begin{tabular}[c]{@{}c@{}}$\triangle$ LM + Llama2 Doc APE\\ Continuous SW\end{tabular} & 31.85*                & 58.9*                 & 0.8298                 & 0.69*         & 0.72         & 0.71          & 0.68*         & 0.81*         & 0.74*         & 0.62*         & 0.64         & 0.63          \\ \midrule
\begin{tabular}[c]{@{}c@{}}$\triangle$ LM + Llama2 Doc APE\\ Gold Target Context\end{tabular} & \textbf{34.59}        & \textbf{59.6}         & \textbf{0.8347}        & \textbf{0.73} & \textbf{0.8} & \textbf{0.76} & \textbf{0.77} & \textbf{0.78} & \textbf{0.77} & \textbf{0.69} & \textbf{0.77} & \textbf{0.73} \\ \bottomrule
\end{tabular}
}
\caption{Comparing our Document Level APE with Llama2 with sentence level APE and conventional approaches. We use chunk-based decoding unless it is explicitly mentioned for Doc2Doc models. We report BLEU, ChrF2 and COMET scores for sentence level evaluation and MuDA tagger scores for document level. The best score in each metric is highlighted in \textbf{bold}. We also compare APE models without gold target context in isolation and append * for the best score in each metric.}
\label{tab:llmdoc2doc}
\end{table*}

APE improves COMET score on ACL for 3.3B NLLB model (0.5 gain) but hurts lexical metrics, suggesting it rephrases outputs while maintaining quality. However, APE harms 54B NLLB translations, likely due to difficulty finding errors in such strong models and adapter training focused on lower-quality hypotheses.


\section{Llama2 as Document-Level Post Editors}

Another motivation for our approach is to exploit LLMs ability to process long sequences for Doc APE. In this section, we evaluate and analyze the performance of our Doc APE model in detail.

To gain insights on whether the Doc APE with Llama2 is beneficial, we compare it against several models. Apart from the previously mentioned sentence-level models such as \textit{$\triangle$ LM} and \textit{Llama2 + LoRA}, we also extend them to the document-level by concatenating sentences \citep{tiedemann-scherrer-2017-neural} (\textit{Doc2Doc}). Furthermore, we evaluate different decoding strategies and report both sentence and document-level metrics in Table \ref{tab:llmdoc2doc}.

After tuning on the dev data, we set the LLama2 maximum chunk token sizes as 1024 for training and 256 for inference (See Figure \ref{fig:data} for more information). This ensured at least 5 preceding sentences for most data, which we found to be reasonable given the computation requirements with large inputs. For $\triangle$ LM Doc2Doc, we use a smaller chunk size (128 tokens) due to its limited capacity.


\textbf{Concatenating Sentences for Doc2Doc Proves Insufficient}: Our analysis reveals that models fine-tuned with $\triangle$ LM and Llama2 separately at the document level exhibit subpar performance when compared to the sentence-level $\triangle$ LM across all considered metrics. This limitation likely stems from the scarcity of document-level parallel data, a common occurrence, particularly in the context of low-resource languages. This highlights the inadequacy of concatenation as a standalone approach in practical use cases.

\textbf{Navigating the Trade-off between Sentence and Document APE}: Doc APE models outperform sentence-level on BLEU/ChrF2 (despite slight COMET dip of $0.3$ between Sent and Doc APE), showing promise for document translation. For pronouns and formality, the Doc APE models leverage context effectively by achieving the best F1 scores of $0.75$ and $0.74$. However, it is still unclear why the COMET score of Doc APE model is worse while we observe improvements in all other metrics.

\textbf{Impact of Decoding Strategy}: Doc APE's different decoding strategies (chunking, windowing) show no clear winner in the sentence or document-level metrics. Batched sliding window, though computationally expensive, offer no significant advantage. Thus, a continuous sliding window or chunking may be preferred for efficiency. However, further research across domains and languages is crucial for a comprehensive understanding of Doc2Doc decoding strategies.

\subsection{Incorporating Target Context during Manual Post Editing}
Until now, we have focused on APE and assumed there is no human feedback. However, in the case of manual PE, we force decode the previous target sentences as the manually corrected target context and condition the model to generate the translation of the current source sentence. We denote this as \textit{$\triangle$ LM + Llama2 Doc APE Gold Target Context} in Table \ref{tab:llmdoc2doc}.

By feeding gold target sentences as context to Doc APE, we achieve substantial gains across metrics: +4.14 BLEU, +2.6 ChrF2, +0.268 COMET, compared to sentence-level $\triangle$ LM. This not only validates Doc APE's ability to leverage context but also suggests the potential for reducing manual edits in PE, leading to cost savings.



\subsection{Disambiguating Pronouns with Doc APE}

We also report scores on the ContraPro test set \citep{muller-etal-2018-large}. This is a benchmark designed to assess the disambiguation of pronouns, specifically "Er" (masculine)," "Sie," (feminine) and "Es" (neutral) when translating "It" from English to German. We evaluate on two setups following \citet{post2023escaping}. For contrastive, we force-decode the target context and determine which pronoun is most likely based on the log-likelihood. In the case of generative, we directly translate the full source paragraph and extract the last sentence to check if it contains the correct pronoun.

\begin{table}[!ht]
\resizebox{\columnwidth}{!}{
\begin{tabular}{@{}c|ccc@{}}
\toprule
             & Cxt Size & Contra/Gen (\%)  \\ \midrule
\citet{post2023escaping}         &       10       &     77.9/\textbf{70.5}             \\
\citet{lupo-etal-2023-encoding}         &     4         &      82.54/\_             \\
$\triangle$ LM + Llama2 Doc APE  &   2       & 87.7/68.0       &            \\
$\triangle$ LM + Llama2 Doc APE & 4            & \textbf{88.7}/69.7       &            \\\bottomrule
\end{tabular}
}
\caption{Contrastive and Generative accuracy on the ContraPro English $\rightarrow$ German Test Set. Results for Sent APE and additional configurations in Table \ref{tab:contrapro_full}}.
\label{tab:contrapro}
\end{table}

We find that our document-level APE model achieves state-of-the-art accuracy 88.7\% in choosing the right pronoun. This can be attributed to LLMs pre-training in long texts. For generative accuracy, we are very comparable to \citet{post2023escaping} with fine-tuning only on TED talks. Moreover, this shows that when target context is made available, LLMs seem to better exploit them and are ideally suited for document-level tasks.


\section{Related Work}
\textbf{Document NMT}: Conventional approaches in Doc-NMT rely on a straightforward concatenation technique \citep{tiedemann-scherrer-2017-neural,agrawal-etal-2018-contextual,post2023escaping}. Several works also explored complicated adaptations to transformer architectures, such as the inclusion of additional context encoders \citep{jean2017does,voita-etal-2018-context}, adjustments to positional information \citep{bao-etal-2021-g,li2023p}, and the application of data augmentation strategies \citep{sun-etal-2022-rethinking}, among others. Similar to our work is \citet{voita-etal-2019-context}, where sentence-level translations are refined to create a coherent document but without considering the source context. 


\textbf{LLM for MT}: LLMs are currently being explored for MT given their success in several tasks. These techniques were mainly facilitated by ICL \citep{brown2020language} at sentence-level \citep{zhang2023prompting,vilar2022prompting} or document-level \citep{hendy2023good,wang2023document}. Similar to our work, the other line of direction is integrating translation memories \citep{mu-etal-2023-augmenting,moslem-etal-2023-adaptive} or correcting NMT system outputs in the prompt \citep{raunak2023leveraging,chen2023iterative}. It is worth noting that our work sets itself apart from these approaches by leveraging efficient LoRA and enabling the effective fusion of NMT with LLMs at both the sentence and document-levels. 

\textbf{Online Learning for NMT:} Integrating human feedback for MT was explored in both statistical MT \citep{formiga2015leveraging,logacheva2017human}. Many methods perform additional training steps using the feedback and alter the MT model at run-time \citep{turchi2017continuous,kothur-etal-2018-document}. Few works explored integrating retrieval and cache mechanisms to avoid further fine-tuning \citep{gu2018search,shang2021guiding,wang2022non}. Our approach incorporates human feedback as context and does not need any changes.

\section{Conclusion}

Our work highlights LLMs' potential for APE, significantly boosting NMT at both sentence and document levels. We showed that it enables modularity, deeper text understanding, and document-level quality boosted by LLMs' massive pretraining.

For future work, we consider several research avenues. These include training the adapters on substantially larger volumes of document-level parallel data, assessing various open-source LLMs, and conducting similar experiments with low-resource languages and domain. 

\section{Limitations}

The main disadvantage of the proposed cascaded system is the latency to generate a translation. From Table \ref{tab:llmdoc2doc}, we find the $\triangle$ LM performance is worse but comparable to the APE approaches with LLM. However, $\triangle$ LM can produce translations with significantly shorter latency compared to LLMs. Therefore, integrating techniques from quality estimation to decide when to perform APE may be helpful to overcome this limitation.

The other drawback of the cascaded approach is that it does not simulate a deep fusion. The LLM can make mistakes even when the NMT is highly confident and correct for a given translation. However, fusing them is not trivial due to the models having different vocabularies.

Finally, we also like to mention that we performed experiments for only English to German direction which was highly present during LLMs pretraining. The benefits of APE should also be validated for low-resource languages for generalizability where the monolingual data of such languages may be significantly less in the LLM pretraining.
\bibliography{anthology,custom}

\appendix

\section{Prompt Templates}
\label{sec:prompt}

In this Section, we provide our prompt templates for our experiments

\subsection{Prompt: LLM In-Context-Learning}
\label{prompticl}

Below is the prompt template for our few-shot ICL experiments. In this example, we perform 2-shot. Given, the two previous examples are either selected randomly or nearest in the embedding space for the source sentences. 
\begin{verbatim}
    ### INSTRUCTION:
    Translate the input from
    English to German.

    ###Input: [SRC1]
    ####Response: [TGT1]

    ###Input: [SRC2]
    ####Response: [TGT2]
    
    ###Input: [SRC3]
    ####Response: 
\end{verbatim}

\subsection{Prompt: LLM Adapter}
\label{promptadapter}
We use the following template when adapting the LLM for sentence-level translation. Note that it is different from the ICL in Section \ref{prompticl}. However, we experimented with the below prompt for ICL and found similar results. We do not again perform the experiments for all configurations due to the computational load. 
\begin{verbatim}
    [INST] <<SYS>>\nYou are a
    professional translator
    from English to German. 
    
    The output should only be the
    translation in one line.<</SYS>>
    
     English: [SRC]
     [/INST]
     German: 
\end{verbatim}

\subsection{Prompt: LLM ZeroShot PE}
\label{promptzerope}

Since this scenario is zero-shot, we provide more instructions so that is much easier for the model to understand the task. Since we found it was generating explanations and notes even when explicitly mentioned not to, we ask to always end the answer with "\#\#\#". Later, we use it as a separator for parsing the output. We provide the prompt template below

\begin{verbatim}
    [INST] <<SYS>>You are a post-editor.
    You improve translations from English
    to German using the English source and
    German translation. Do not provide any
    explanation or correction.
    The translation should end with
    ### in new line
    <</SYS>>
    English: [SRC]
    German Translation: [HYP]
    [/INST]
    Post-Edited Translation:
\end{verbatim}

\subsection{Prompt: LLM LoRA Sentence APE}
\label{promptsentpe}

In this case, we decrease the prompt size as now we perform fine-tuning and instructions are not necessary. Furthermore, it will lead to less memory consumption as the sequences are much shorter.

\begin{verbatim}
    English: [SRC]
    German Translation: [HYP]
    Post-Edited Translation: [REF]  
\end{verbatim}

\subsection{Prompt: LLM LoRA Document APE}
\label{promptdocpe}

We format the prompt similarly to the sentence level. The only difference is that now we have sentences separated by "<SS>" token in the document.

\begin{verbatim}
    English: [SRC1] <SS> [SRC2] <SS> [SRC3]
    German Translation: [HYP1] <SS> [HYP2] 
                        <SS> [HYP3]
    Post-Edited Translation: [REF1] <SS> 
                        [REF2] <SS> [REF3]  
\end{verbatim}

\section{LLM In-Context-Learning: Zero to 5-shot}
\label{sec:llmicl}

Please refer to Table \ref{tab:iclrandom} for results on ICL with random selection and Table \ref{tab:iclfaiss}. The results reported are on the Must-C v3 test set using LLama2.

\begin{table}[!ht]
\begin{tabular}{@{}cccc@{}}
\toprule
Shot Size           & BLEU  & ChrF & COMET  \\ \midrule
Sent-level $\triangle$ LM & 30.45 & 57.0 & 0.8179 \\
0                   & 20.47 & 48.3 & 0.7592 \\
1                   & 20.73 & 48.8 & 0.7697 \\
2                   & 21.53 & 50.0 & 0.7795 \\
3                   & 20.34 & 50.1 & 0.7685 \\
4                   & 19.87 & 50.0 & 0.7609 \\
5                   & 20.33 & 50.5 & 0.7658 \\ \bottomrule
\end{tabular}
\caption{In-Context-Learning with Llama2 using Random Selection Strategy}
\label{tab:iclrandom}
\end{table}

\begin{table}[!ht]
\begin{tabular}{@{}cccc@{}}
\toprule
Shot Size           & BLEU  & ChrF & COMET  \\ \midrule
Sent-level $\triangle$ LM & 30.45 & 57.0 & 0.8179 \\
0                   & 20.47 & 48.3 & 0.7592 \\
1                   & 21.16 & 49.8 & 0.7755 \\
2                   & 21.13 & 50.2 & 0.7724 \\
3                   & 19.61 & 49.8 & 0.7593 \\
4                   & 18.82 & 49.9 & 0.7531 \\
5                   & 18.51 & 49.9 & 0.7402 \\ \bottomrule
\end{tabular}
\caption{In-Context-Learning with Llama2 using FAISS Selection Strategy}
\label{tab:iclfaiss}
\end{table}

\section{Training Details}

\subsection{Llama2 Experiments}
\label{sec:train_details}

We use the transformers library \citep{wolf-etal-2020-transformers} for training and inference with Llama2. While training the adapters, we set the hyper-parameters to rank 8, alpha 32, dropout 0.1, and bias as \textit{'LoRA\_only'}. Following \citet{dettmers2023qLoRA} to make the model robust to LoRA hyper-parameters, we adapt on all layers. The modules we add to the adapter include \textit{q\_proj,k\_proj,v\_proj,gate\_proj,up\_proj and down\_proj}. We set a batch size for each device to 32 initially and enable \textit{auto\_find\_batch\_size} to \textit{True} on 4 NVIDIA RTX A6000 GPU's. To simulate a larger batch size, we set \textit{gradient\_accumulation\_steps} to 20. We use a \textit{learning\_rate} of $2e-5$. The other parameters are set to default. We train for $3$ epochs and select the model with the best validation loss. During inference, we use beam search with a \textit{num\_beams} set as 3 as we find it to be reasonable given the computation and performance.

\subsection{DeltaLM Experiments}
\label{sec:train_details_deltalm}
We use the fairseq library \citep{ott-etal-2019-fairseq} for our experiments with $\triangle$ LM. During training, we use cross-entropy loss with label smoothing set to $0.1$. We set a learning rate of $0.0001$ with Adam optimizer, betas $(0.9,0.98)$ and the initial learning rate to $1e-7$. We set both dropout and attention dropout to $0.1$. We use a batch size of $2000$ max tokens and perform gradient accumulation for 3 steps. We train until the validation loss increases after 5 consecutive interval steps that are set to $4500$ steps (Roughly $1/3$ of epoch). During inference, we do beam size with the number of beams set to 5. The other parameters not mentioned are set to default.

\section{ContraPro Scores for Sentence and Document APE}

\begin{table}[!hbt]
\resizebox{\columnwidth}{!}{
\begin{tabular}{@{}c|ccc@{}}
\toprule
             & Cxt Size & Contra/Gen (\%)  \\ \midrule
\citet{post2023escaping}         &       10       &     77.9/70.5             \\
\citet{lupo-etal-2023-encoding}         &     4         &      82.54/\_             \\
$\triangle$ LM + Llama2 Sent APE      & 0            & 60.0/\_         &            \\
$\triangle$ LM + Llama2 Sent APE    & 2            & 85.8/\_       &            \\
$\triangle$ LM + Llama2 Doc APE & 0            & 50.9/\_       &            \\
$\triangle$ LM + Llama2 Doc APE  &   2       & 87.7/68.0       &            \\
$\triangle$ LM + Llama2 Doc APE & 4            & \textbf{88.7}/69.7       &            \\\bottomrule
\end{tabular}
}
\caption{Comparing Sentence and Document APE Accuracy on the ContraPro English $\rightarrow$ German Test Set. For generative results, we only report on sentences from 1 to 10 using th evaluation script from \citet{post2023escaping}.}
\label{tab:contrapro_full}
\end{table}

\begin{table*}[ht]
\centering
\resizebox{2\columnwidth}{!}{
\begin{tabular}{@{}l|l@{}}
Source                    & \makecell{This is a sentence in Spanish: Las prendas bestsellers se estampan \\  con motivos fLoRAles, animal print o retales tipo patchwork.}   \\ \midrule
Reference                 & \makecell{Dies ist ein Satz auf Spanisch: Las prendas bestsellers se estampan \\ con motivos fLoRAles, animal print o retales tipo patchwork.}   \\ \midrule
$\triangle$ LM Hypothesis & \makecell{Das ist ein Satz auf Spanisch: Die Bestsellers se multidisciplinan \\ conão fLoRAles, Tierdruck oder Reliefs ol Flitterwerk.}   \\ \midrule
Post-Edited with Llama2      & \makecell{Das ist ein Satz auf Spanisch: Las prendas bestsellers se estampan \\ con motivos fLoRAles, animal print o retales tipo patchwork.}
\end{tabular}
}
\caption{Example from the ACL dev set taken from Talk id: 268 and Sentence 26. The $\triangle$ LM translates everything into German including the Spanish phrase that needs to be retained in the original language. However, after APE, Llama2 does not translate the Spanish Phrase as it was also not translated in the source sentence.}
\label{tab:pesentexample}
\end{table*}

\begin{figure*}
     \centering
     \begin{subfigure}[b]{0.4\textwidth}
         \centering
         \includegraphics[width=\textwidth]{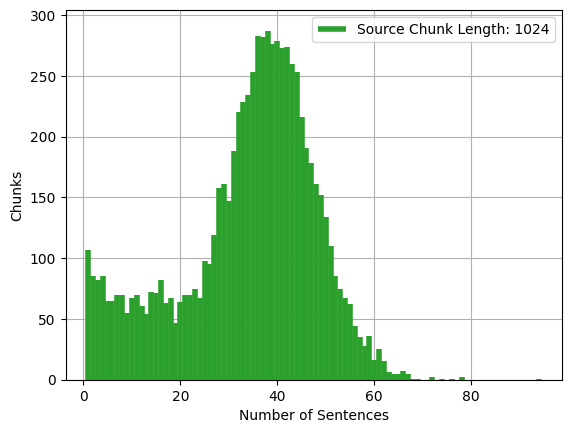}
         \label{fig:y equals x}
     \end{subfigure}
     \hfill
     \begin{subfigure}[b]{0.4\textwidth}
         \centering
         \includegraphics[width=\textwidth]{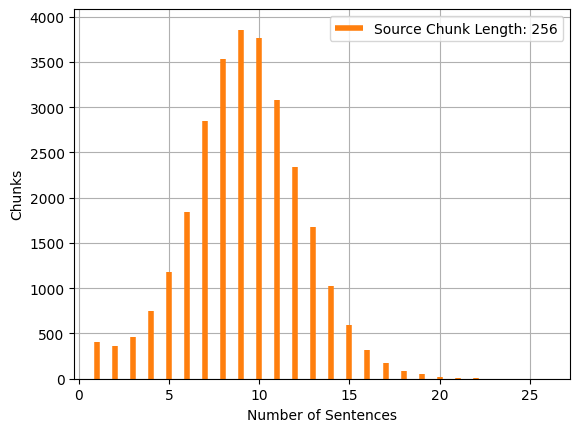}
         \label{fig:three sin x}
     \end{subfigure}
     \hfill
     \caption{Number of sentences in a document with chunk sizes 1024 and 256.}
     \label{fig:data}
\end{figure*}

\end{document}